\documentclass{article}

\usepackage[nonatbib,final]{neurips_2022}

\usepackage[utf8]{inputenc} 
\usepackage[T1]{fontenc}    
\usepackage{hyperref}       
\usepackage{url}            
\usepackage{booktabs}       
\usepackage{amsfonts}       
\usepackage{nicefrac}       
\usepackage{microtype}      
\usepackage{xcolor}         
\usepackage{cite}
\usepackage{graphicx}
\usepackage{wrapfig}
\usepackage{subcaption}
\usepackage{comment}

\title{A Self-Supervised Approach to Reconstruction in Sparse X-Ray Computed Tomography}

\author{Rey Mendoza\thanks{
Equal contribution. Correspondence to
\texttt{vidyag@berkeley.edu}.
}~ $^1$, Minh Nguyen$^{*1}$, 
Judith Weng Zhu$^{*1}$,\\
\textbf{Vincent Dumont$^{2}$, 
Talita Perciano$^{2}$, 
Juliane Mueller$^{2,3}$, 
Vidya Ganapati$^{1,2}$}\\
    $^1$ Swarthmore College, $^2$ Lawrence Berkeley National Laboratory, \\ $^3$ National Renewable Energy Laboratory
  }

\begin{document}

\maketitle

\vspace{-1.0em}
\begin{abstract}

Computed tomography has propelled scientific advances in fields from biology to materials science. This technology allows for the elucidation of 3-dimensional internal structure by the attenuation of x-rays through an object at different rotations relative to the beam. By imaging 2-dimensional projections, a 3-dimensional object can be reconstructed through a computational algorithm. Imaging at a greater number of rotation angles allows for improved reconstruction. However, taking more measurements increases the x-ray dose and may cause sample damage. Deep neural networks have been used to transform sparse 2-D projection measurements to a 3-D reconstruction by training on a dataset of known similar objects. However, obtaining high-quality object reconstructions for the training dataset requires high x-ray dose measurements that can destroy or alter the specimen before imaging is complete. This becomes a chicken-and-egg problem: high-quality reconstructions cannot be generated without deep learning, and the deep neural network cannot be learned without the reconstructions. This work develops and validates a self-supervised probabilistic deep learning technique, the physics-informed variational autoencoder, to solve this problem. A dataset consisting solely of sparse projection measurements from each object is used to jointly reconstruct all objects of the set. This approach has the potential to allow visualization of fragile samples with x-ray computed tomography. We release our code for reproducing our results at: \url{https://github.com/vganapati/CT_PVAE}.

\end{abstract}

\section{Introduction \& Related Work}

\subsection{Computed Tomography}

The power of x-rays to penetrate through many materials has allowed for advances in scientific understanding. In particular, computed tomography has been used in fields such as biology \cite{wise_micro-computed_2013}, medicine \cite{rubin_computed_2014}, materials science \cite{garcea_x-ray_2018}, and geoscience \cite{cnudde_high-resolution_2013} to elucidate the internal structure of samples. Computed tomography works by measuring the attenuation of an x-ray beam through an object at different rotations relative to the beam. From the projection images created by parallel beams at each rotation, a 3-dimensional object can be reconstructed with a computational algorithm. Imaging at many rotation angles allows for improved reconstruction, however, fragile samples can only be imaged under limited x-ray dose before damage and structural changes occur. 
 
\subsection{Reconstruction Algorithms}

A 3-dimensional object can be reconstructed from projection images through direct algorithms such as filtered backpropagation or iterative algorithms \cite{pelt_improving_2018, marchesini_sparse_2020}. In recent years, algorithms using deep learning have come into prominence \cite{parkinson_machine_2017, pelt_improving_2018, ayyagari_image_2018, huang_investigations_2018, bazrafkan_deep_2019, fu_hierarchical_2019, he_radon_2019}. In these approaches, the parameters of the deep neural network are found by training on large datasets of known input-output pairs. The computational burden of training is high and a training dataset is required, which can be generated through high-quality experimental data and traditional algorithms. However, once trained, a relatively quick forward pass through the network can generate the reconstruction \cite{pelt_improving_2018}. Deep learning has also shown the potential to improve upon conventional methods when there are a limited number of projection images, as a prior is learned from the data and embedded into the neural network parameters \cite{parkinson_machine_2017, pelt_improving_2018}.

\subsection{Deep Learning without Ground Truth}

While deep learning methods for computed tomography are promising, these approaches require a training dataset of reconstructions. High-quality reconstructions need measurements at many rotation angles, and this high radiation dose may destroy or alter fragile specimens during the data acquisition process. A chicken-and-egg problem presents itself: the deep learning algorithm needs reconstructions, and the reconstructions need the deep learning algorithm.
 
 An approach to solve this problem involves using training datasets consisting only of sparse angle measurements for every object of the set. The intuition is that these sparse measurements taken together can provide enough information to infer a prior distribution of the objects. One method is to train generative adversarial networks to create high-quality reconstructions by rewarding reconstructions that have corresponding sparse measurements that lie in the training distribution of sparse measurements \cite{kabkab_task-aware_2018, bora2018ambientgan, kuanar_low_2019, Cole_2021_ICCV, xu_fast_2019}.  Another method is to train directly with low-dose measurements, penalizing reconstructions by the discrepancy between simulated low-dose and true low-dose measurements \cite{tamir_unsupervised_2020, gan_image_2020, Zhussip_2019_CVPR, liu_rare_2020}. Finally, other works use multiple low-dose measurements of each object in training \cite{xia_training_2019, lehtinen_noise2noise_2018}, though this may be unsuitable for fragile samples that cannot withstand repeated measurement. 

\subsection{Probabilistic Formulation}

In all deep learning methods without ground truth surveyed, the reconstruction is given as a point estimate. In this work, we aim to find the posterior probability distribution $P(O|M)$, where $O$ is the object being reconstructed and $M = [M_1, M_2, ..., M_n]$ are the measurements, allowing the uncertainty of the prediction to be modeled. We aim to lower the total number of measurements $n$ to minimize the x-ray dose. We assume that we have a set of $m$ objects $\{O_1, O_2, ..., O_m\}$, sampled from some distribution $P(O)$, and we aim to reconstruct all objects in the set. For each of the $m$ objects, we are allowed $n$ measurements. Each set of measurements for an object $j$, $M_j = [M_{j1}, M_{j2}, ..., M_{jn}]$ is obtained with chosen rotation angles $p_j = [p_{j1}, p_{j2}, ..., p_{jn}]$. We assume that the forward model physics $P(M | O; p) = P(M | O)$ is known; in computed tomography the forward model is the Radon transform with Poisson noise. For every object $O$, we aim to find the posterior distribution $P(O | M) = \frac{P(M | O) P(O)}{P(M)}$. This work tackles the following problems that arise in finding the posterior: (1) construction of the prior $P(O)$ with no directly observed $O$, only measurements $M$ on each object of the set, (2) calculating $P(O | M)$ in a tractable manner. 

\section{Methods}

\subsection{Physics-Informed Variational Autoencoder}

We create a framework for posterior estimation that is inspired by the mathematics of the variational autoencoder \cite{kingma2014autoencoding, doersch2021tutorial}. In a variational autoencoder, the goal is to learn how to generate new examples, sampled from the same underlying probability distribution as a training dataset of objects. To accomplish this task, a latent random variable $z$ is created that describes the space on a lower-dimensional manifold. A deep neural network defines a function (the ``decoder'') from a sample of $z$ to a probability distribution $P(O|z)$. The parameters of the deep neural network are optimized to maximize the probability of generating the objects of the training dataset.

%

In this work, we aim to find the posterior probability distribution $P(O|M)$, where $O$ is the object being reconstructed and $M$ are the projection measurements. In our case, we only have a dataset of noisy measurements $M$ and no ground truth objects $O$. However, the forward model, $P(M|O)$, is known. Thus, instead of maximizing the probability of generating $O$, we can maximize the probability of generating $M$, a formulation we call the ``physics-informed variational autoencoder.''

\begin{figure}[h]
\vspace{-1.0em}
\centering
\includegraphics[width=1\textwidth]{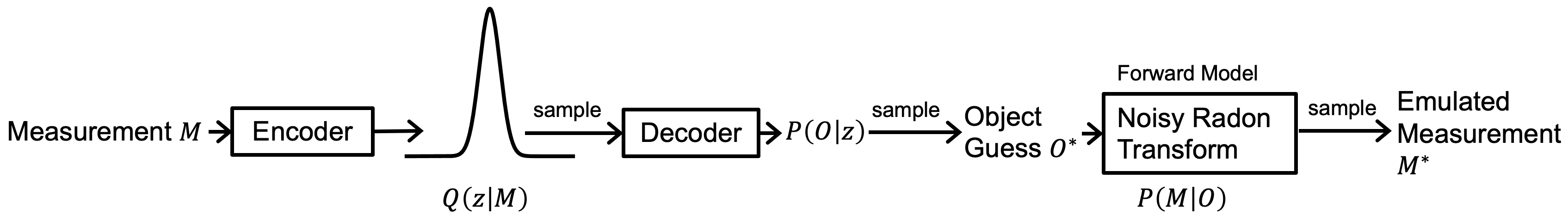}
\vspace{-1.5em}
\caption{The developed physics-informed variational autoencoder.}
\vspace{0.0em}
\label{fig:vae}
\end{figure}

\vspace{0em}

We aim to maximize $P(M)= \int \int P(M|O)P(O|z)P(z) dO dz$. To compute this integral in a computationally tractable manner, we can approximate with sampled values. As in \cite{kingma2014autoencoding, doersch2021tutorial}, for most randomly sampled values of $z$ and $O$, the probability $P(M|O,z)$ is close to zero, causing poor scaling of sampled estimates. Similar to variational autoencoders, our framework solves this problem by estimating the parameters of $P(z|M)$ by processing the measurements $M$ using a function with trainable parameters (called the ``encoder,'' also formulated as a deep neural network). The estimate of $P(z|M)$ is denoted $Q(z|M)$. The Kullback–Leibler divergence between the distributions is given by $D[Q(z|M) || P(z|M)] = E_{z \sim Q}[\log Q(z|M) - \log P(z|M)]$. We also have, by Bayes' Theorem, $\log P(z|M) = \log P(M|z) + \log P(z) - \log P(M)$. Combining the expressions yields:
\[
 \log P(M) - D[Q(z|M) || P(z|M)] = E_{z \sim Q} \left[ \int P(M|O)P(O|z)dO\right] - D\left[\log Q(z|M) || \log P(z) \right]. 
\]
The first term on the right side of this expression can be estimated with sampled values. As Kullback–Leibler divergence is always $\geq 0$ and reaches $0$ when $Q(z|M) = P(z|M)$, maximizing the right side (defined here as the loss) during training causes $P(M)$ to be maximized while forcing $Q(z|M)$ towards $P(z|M)$. In contrast to a conventional variational autoencoder, we do \textit{not} attempt to use this formulation to synthesize arbitrary objects $O$ by sampling $P(z)$ directly. This framework only attempts reconstruction on the training examples themselves. After training, $P(O|M)$ for every object can be sampled by first sampling $Q(z|M)$ then $P(O|z)$; see Fig.~\ref{fig:vae}. Crucially, unlike most data-driven approaches to reconstruction in computational imaging, the developed framework assumes that no ground truth dataset of objects $O$ is available.

In this work, the encoder and decoder of Fig.~\ref{fig:vae} take the same basic shape as a U-Net \cite{ronneberger_u-net_2015}. The input to the encoder is an object reconstruction created with a standard algorithm from the sparse set of projections, using the \textsc{TomoPy} Python package \cite{gursoy_tomopy_2014}, as well as specification of the projection angles used. The skip connections to the decoder are parametrized as Gaussian distributions, and sampled to determine the latent variables $z$. The likelihood distribution $P(M | O)$ is given by the Radon transform with the addition of Poisson noise.

\section{Results \& Discussion}
\label{sec:results}

We prototype and validate the physics-informed variational autoencoder with synthetic datasets. Code to reproduce our results are at \url{https://github.com/vganapati/CT_PVAE}. We utilize 2-D objects and 1-D projections without loss of generality; a 3-D object in x-ray computed tomography is reconstructed as a series of 2-D axial slices.

\subsection{Simple Toy Example}

\begin{wrapfigure}{r}{0.29\textwidth}
  \vspace{-2.25em}
  \begin{center}
    \includegraphics[width=0.29\textwidth]{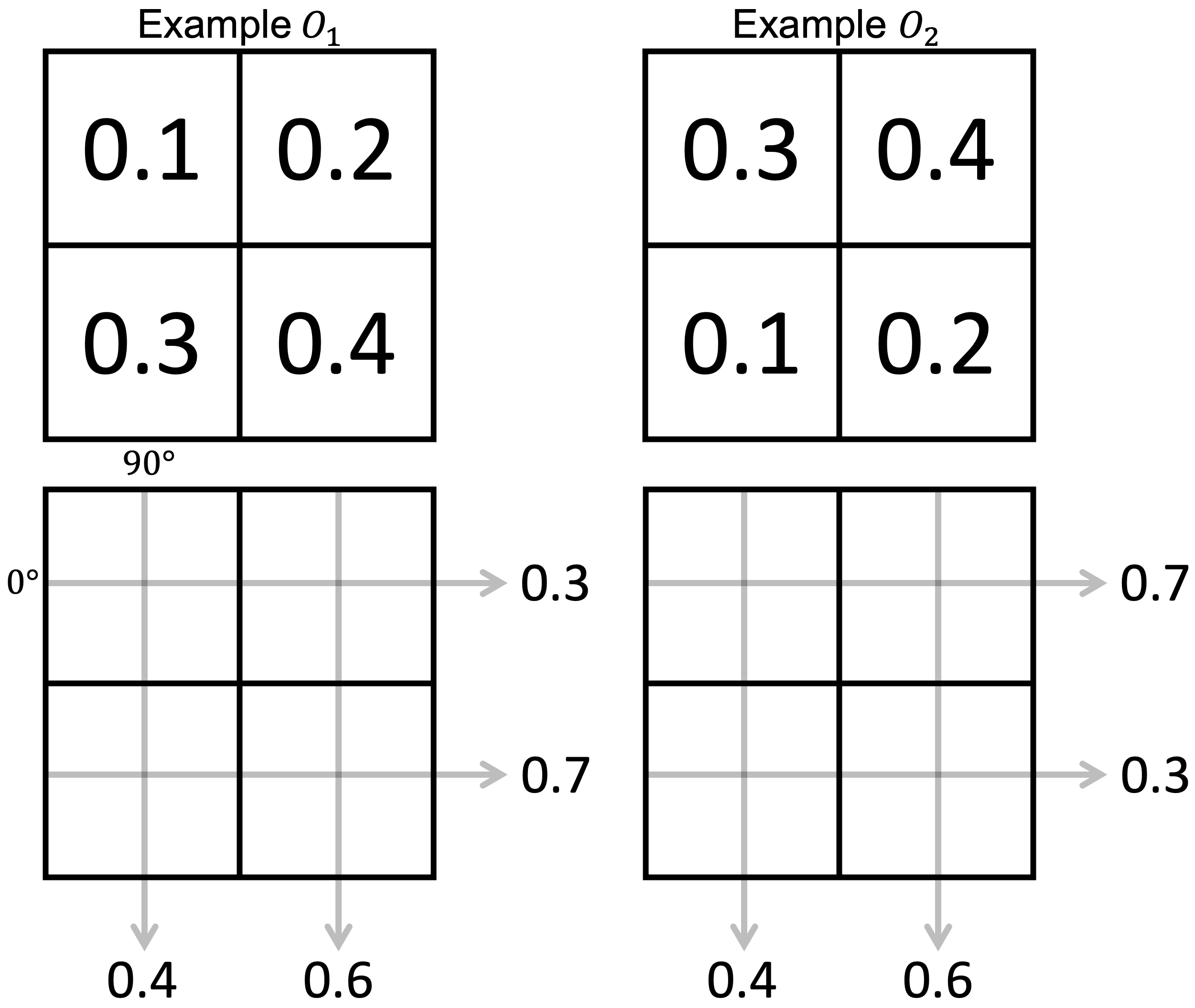}
  \end{center}
    \vspace{-1.0em}
  \caption{Toy dataset consisting of 2 object types.}
  \label{fig:toy}
  \vspace{-1.5em}
\end{wrapfigure}

\begin{figure}[t]
    \vspace{-0.5em}
  \begin{subfigure}[t]{.45\textwidth}
    \centering
    \includegraphics[width=\linewidth]{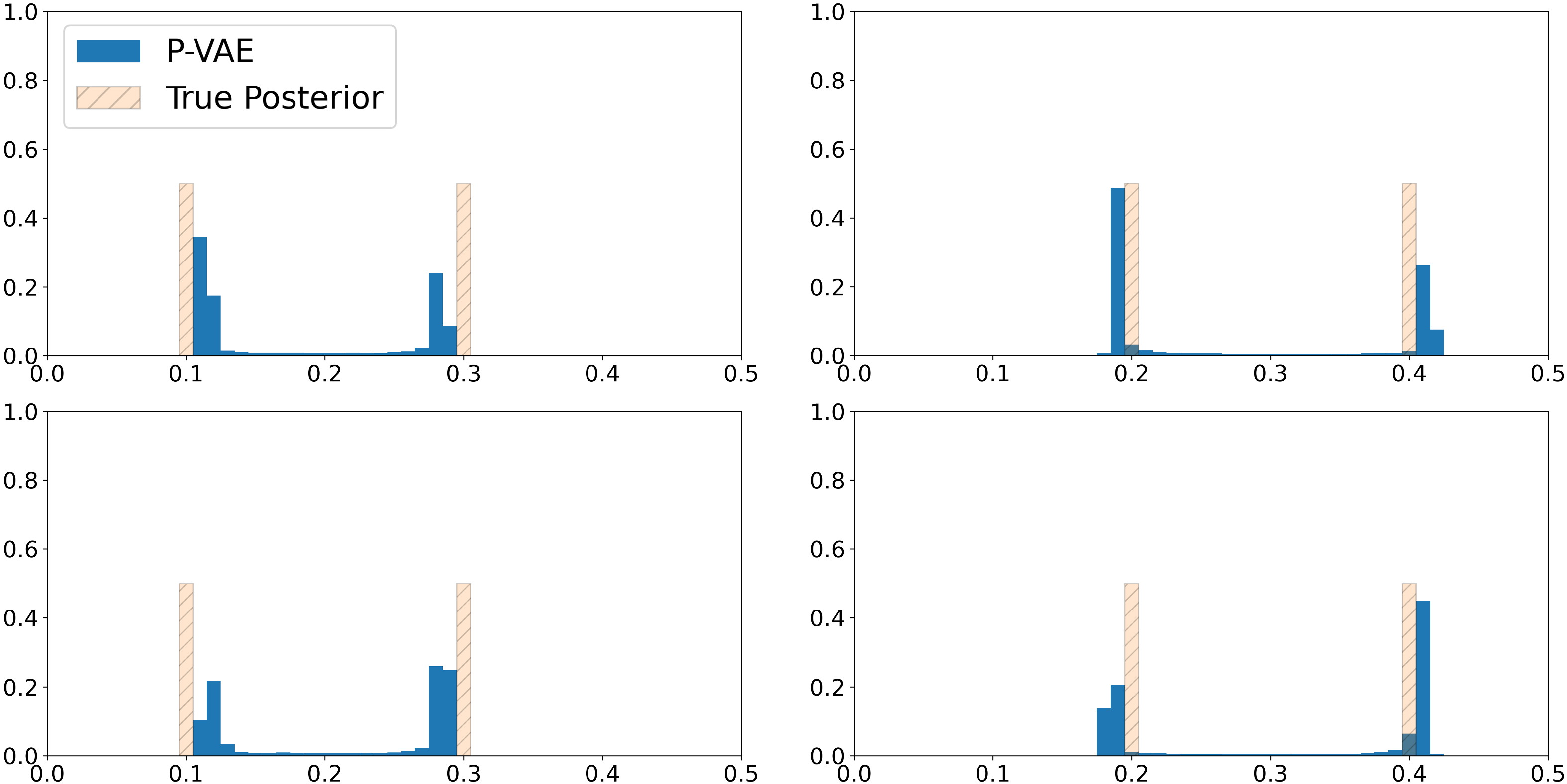}
    \caption{$O_1$ measured at $\frac{\pi}{2}$ radians.}
  \end{subfigure}
  \hfill
  \begin{subfigure}[t]{.45\textwidth}
    \centering
    \includegraphics[width=\linewidth]{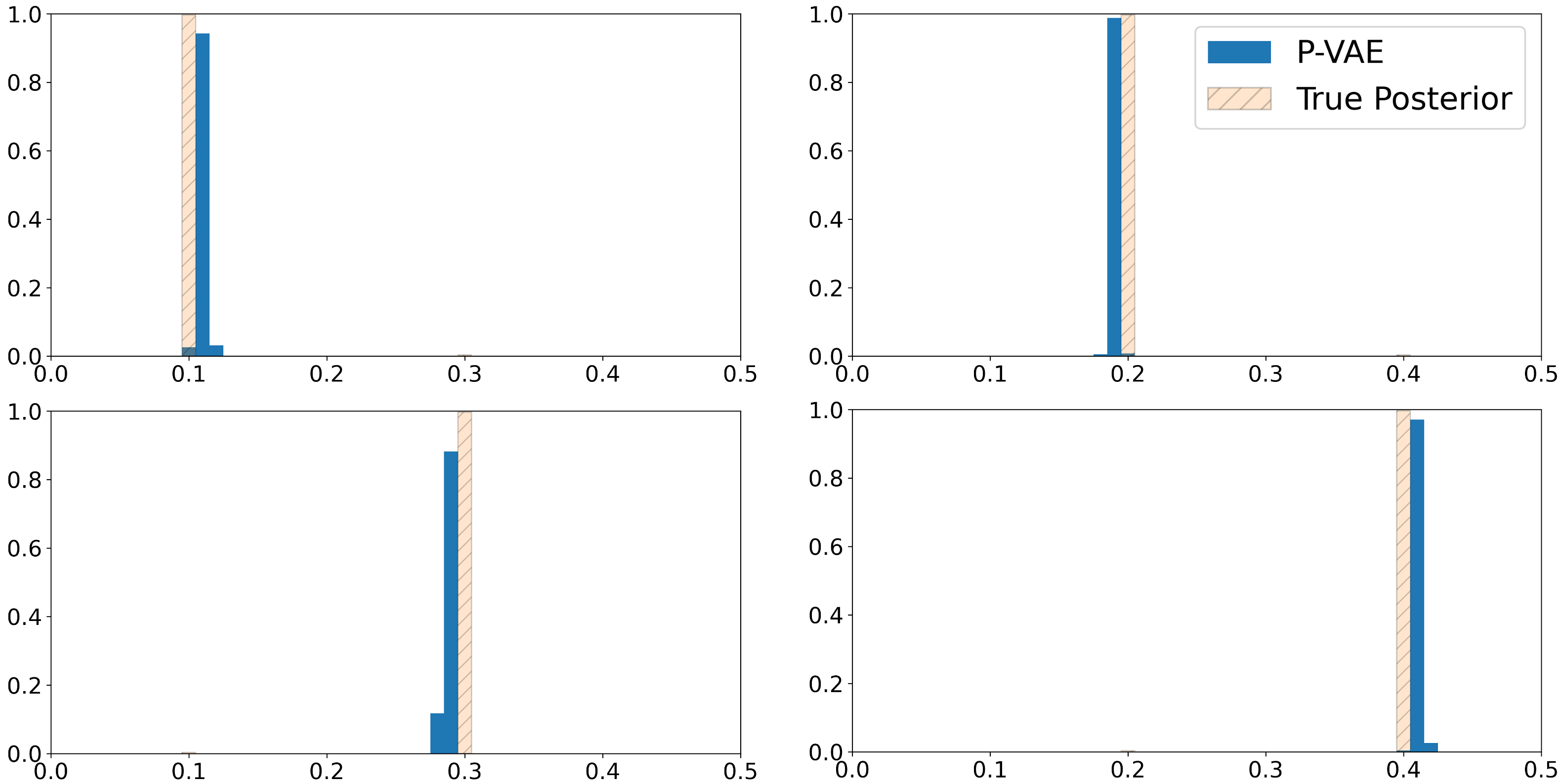}
    \caption{$O_1$ measured at $0$ radians.}
  \end{subfigure}

%
  \caption{Marginal posterior probabilities for each pixel from measurements on the toy dataset; measurements taken on object $O_1$ of the dataset. See Fig.~\ref{fig:toy_results2} in the Appendix for posteriors on $O_2$.}
  \label{fig:toy_results}
      \vspace{-1.0em}
\end{figure}

Our first dataset consists of two unique objects. Each object can be represented by $2 \times 2$ pixels, and projections can be taken at rotation angles of $0$ and $\frac{\pi}{2}$ radians; see Fig.~\ref{fig:toy}. The projection of both the objects at $\frac{\pi}{2}$ is the same, but the projections at $0$ radians are different. We assume that $P(O = O_1) = P(O = O_2) = \frac{1}{2}$. We sample $P(O)$, and then take a noisy measurement $M=[M_1]$ at one rotation angle $p$ (i.e. $n=1$ and $P(p = 0) = P(p= \frac{\pi}{2}) = \frac{1}{2}$). We complete this sampling and measurement procedure $m=1\,024$ times. From all the measurements, and no knowledge of the prior $P(O)$, we aim to determine the posterior $P(O | M)$ for each $M$ by training the physics-informed variational autoencoder. Training on an NVIDIA GeForce GPU takes 47 minutes. Training parameters are available in the released documentation and code. The true posterior is compared with the posterior from $20\,000$ samples of the trained physics-informed variational autoencoder, showing good qualitative agreement; see Fig.~\ref{fig:toy_results}. The marginal posterior for each of the four pixels is compared for four different example measurements. With measurements taken at the rotation angle $\frac{\pi}{2}$, there is approximately equal probability of the pixel values from the two object types. However, with measurements taken at $0$ radians, the posterior is nearly a delta function at the true pixel value. 


%
%
%

\subsection{Synthetic Foam Dataset}


We generate the second dataset of $1\,000$ foam images of $128 \times 128$ pixels through the Python package \textsc{XDesign} \cite{noauthor_xdesign_nodate}, and projections are taken at $180$ equally-spaced angles. The image of the projections is known as the sinogram, and we emulate sparse sinograms by removing $160$ of the projection angles either uniformly or randomly; see Fig.~\ref{fig:foam} in the Appendix for example images and corresponding sinograms of the dataset. We jointly reconstruct all images of this dataset by training the physics-informed variational autoencoder. Training on an NVIDIA Tesla V100 takes 2.2 hours. Training parameters are available in the released code and documentation.

\begin{figure}[h]
    \vspace{-1.0em}
  \begin{subfigure}[t]{.32\textwidth}
    \centering
    \includegraphics[width=\linewidth]{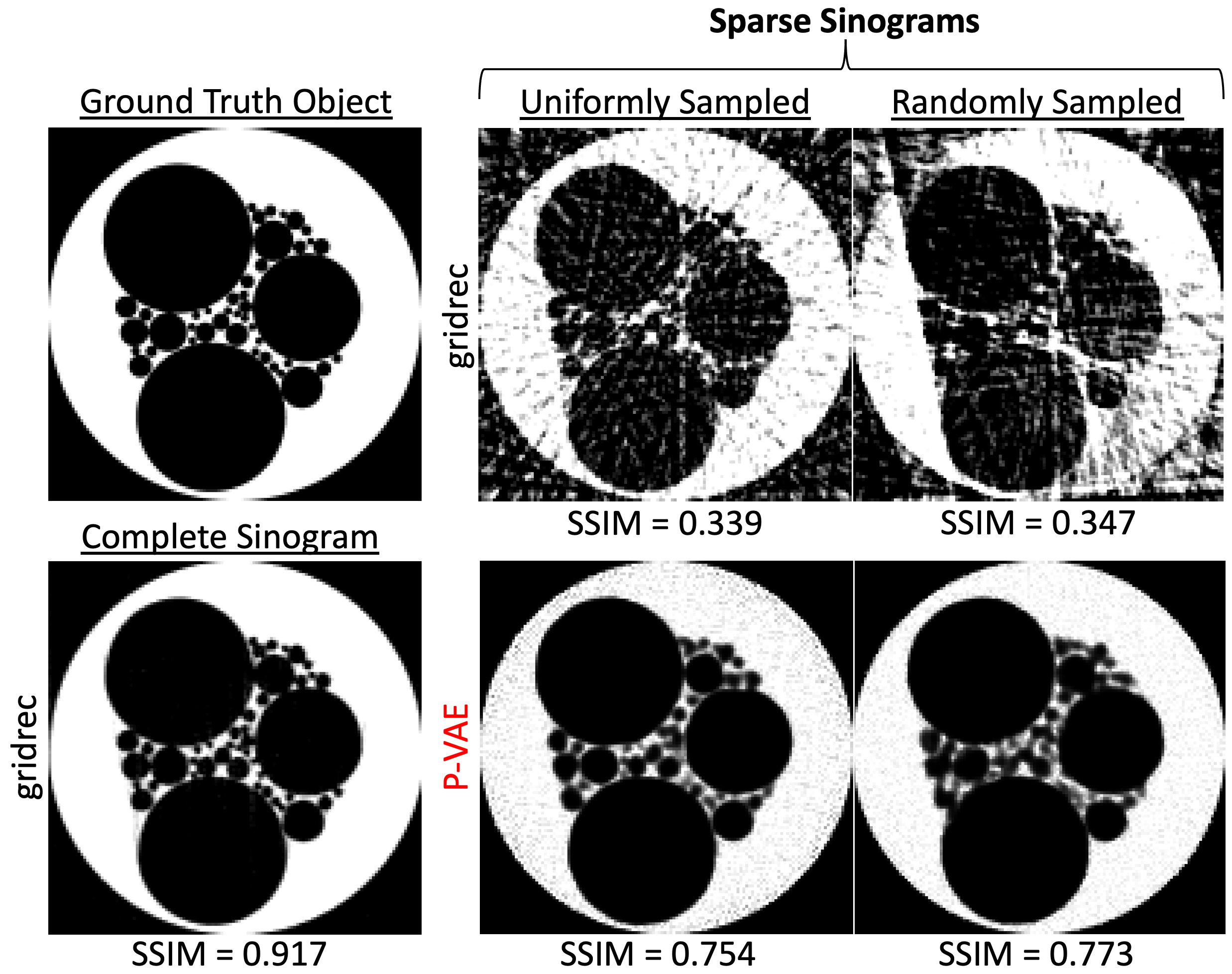}
    \caption{Example 1}
  \end{subfigure}
  \hfill
  \begin{subfigure}[t]{.32\textwidth}
    \centering
    \includegraphics[width=\linewidth]{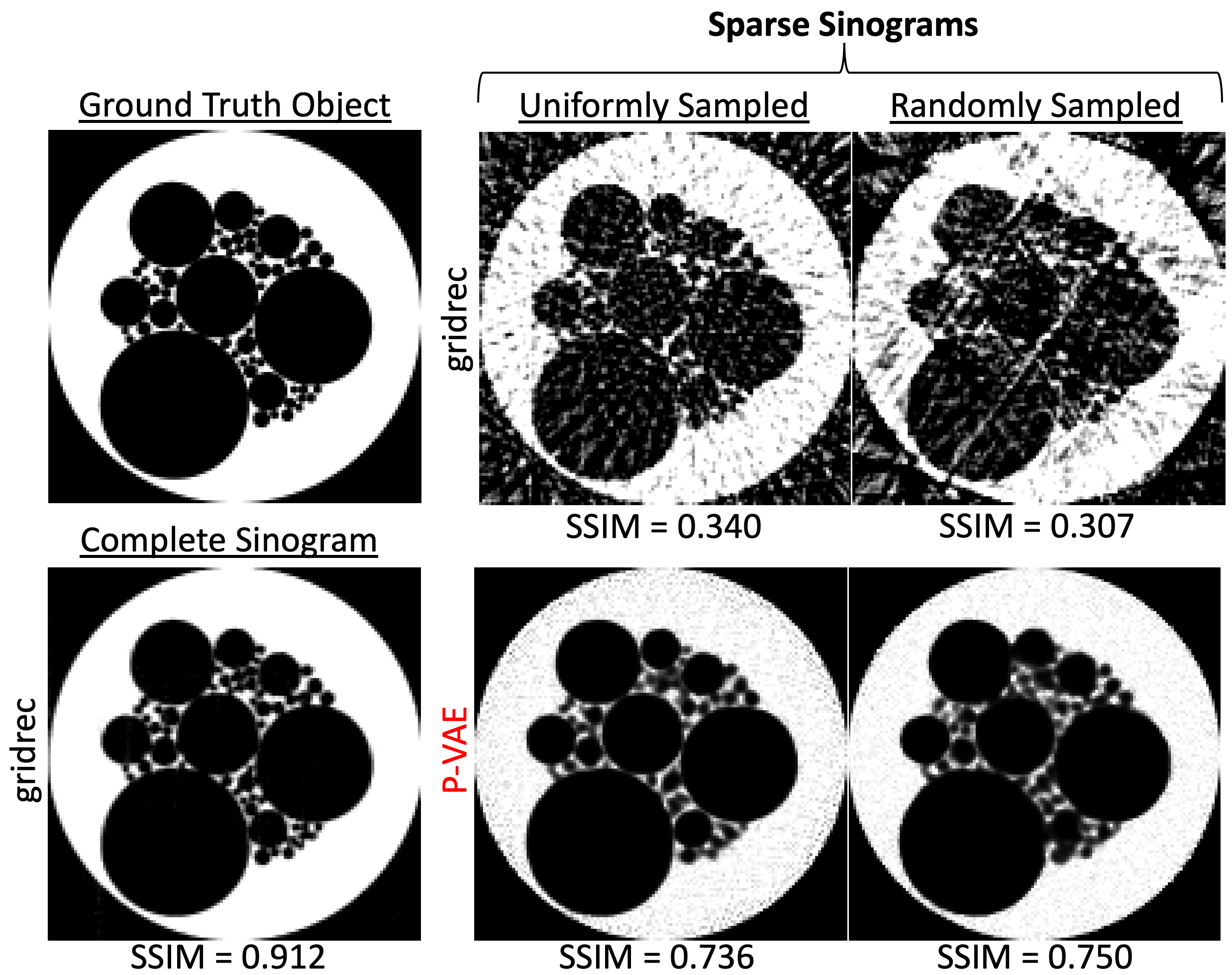}
    \caption{Example 2}
  \end{subfigure}
  \hfill
  \begin{subfigure}[t]{.32\textwidth}
    \centering
    \includegraphics[width=\linewidth]{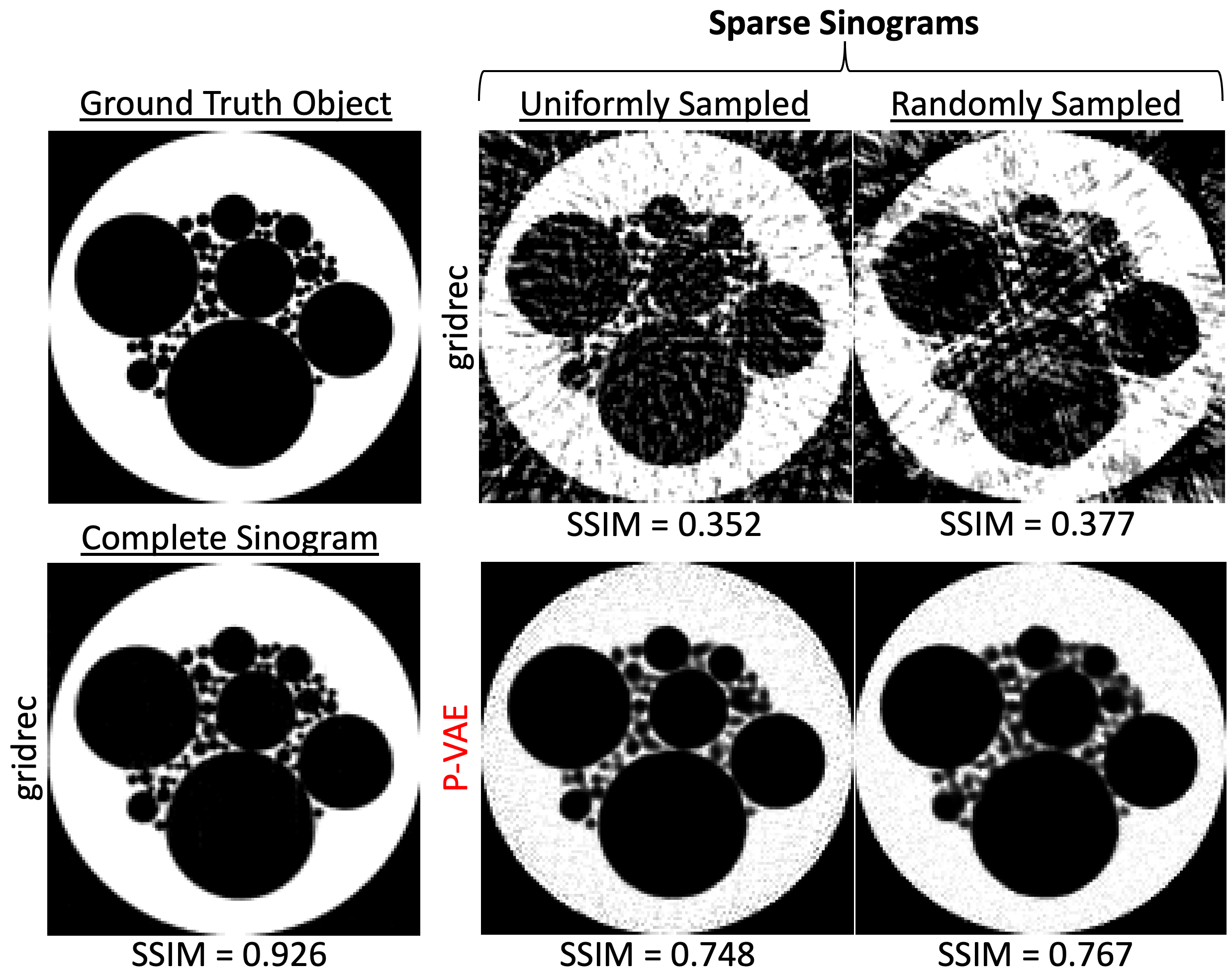}
    \caption{Example 3}
  \end{subfigure}
      \vspace{-0.5em}
  \caption{Foam dataset reconstruction results.}
  \label{fig:foam_results}
      \vspace{-1.0em}
\end{figure}

Fig.~\ref{fig:foam_results} depicts example objects from the synthetic foam dataset, reconstructed with the Fourier grid reconstruction algorithm (\texttt{gridrec}) \cite{dowd_developments_1999}. Reconstructions were performed with two other standard algorithms, the simultaneous algebraic reconstruction technique \cite{gursoy_tomopy_2014} and the Total Variation reconstruction technique \cite{chambolle_first-order_2011}; however \texttt{gridrec} outperformed these techniques on image quality metrics of structural similarity (SSIM), mean-squared error (MSE), and peak signal-to-noise (PSNR). The reconstructions with \texttt{gridrec} on the sparse sinograms are compared with the reconstructions from the physics-informed variational autoencoder (P-VAE); see Fig.~\ref{fig:foam_results}. We observe that the P-VAE improves over \texttt{gridrec} for the sparse sinograms, and there is a slight improvement by randomly selecting the angles in every example over uniform sampling; see Fig.~\ref{fig:foam_aggregate} in the Appendix for results averaged over the entire $1\,000$ object dataset and aggregated over $10$ independent trials.

\section{Conclusions \& Limitations}
\label{sec:conclusion}
We presented a novel data-driven, self-supervised algorithm for image reconstruction in computed tomography and validate on synthetic datasets.  Our method is probabilistic, allowing for quantification of uncertainty in applications, and self-supervised, requiring no ground truth or reference training examples. Though initial results are promising, our work is limited in that we only consider highly synthetic data. Real experimental data may pose additional problems such as model mismatch that need to be considered in the construction of the likelihood $P(M | O)$. In experiments, thousands of rotation angles may be collected with measurements on the order of $100$ GB per object \cite{marchesini_sparse_2020}.  An important direction for future work is high memory management techniques for training. 



\section{Broader Impact}

\label{sec:broader}

Computed tomography uses x-rays to probe the 3-dimensional internal structure of objects, and is widely used in many fields such as medicine, biology, and materials science. However, x-rays can cause sample damage and deformation during the measurement process, limiting the use cases of computed tomography. Much effort has been put into sparse computed tomography, where a limited number of measurements are used to measure and reconstruct an object. This work proposes a novel algorithm for reconstruction in sparse computed tomography, with the potential positive broader impact of making visualization of radiation-damage prone objects possible. However, practitioners should be careful in applying these methods as there are no theoretical guarantees thus far. Application of this method could result in reconstruction artifacts that may have a different appearance than artifacts from conventional methods, evading detection. 


\begin{ack}

This work was supported in part by the U.S. Department of Energy, Office of Science, Office of Workforce Development for Teachers and Scientists (WDTS) under the Visiting Faculty Program (VFP) and the AAUW Research Publication Grant in Engineering, Medicine and Science. 

\end{ack}

\bibliography{neurips_2022}

\section*{Checklist}

\begin{enumerate}

\item For all authors...
\begin{enumerate}
  \item Do the main claims made in the abstract and introduction accurately reflect the paper's contributions and scope?
    \answerYes{}
  \item Did you describe the limitations of your work?
    \answerYes{[See Section \ref{sec:conclusion}.]}
  \item Did you discuss any potential negative societal impacts of your work?
    \answerYes{[See Section \ref{sec:broader}.]}
  \item Have you read the ethics review guidelines and ensured that your paper conforms to them?
    \answerYes{}
\end{enumerate}

\item If you are including theoretical results...
\begin{enumerate}
  \item Did you state the full set of assumptions of all theoretical results?
    \answerNA{}
        \item Did you include complete proofs of all theoretical results?
    \answerNA{}
\end{enumerate}

\item If you ran experiments...
\begin{enumerate}
  \item Did you include the code, data, and instructions needed to reproduce the main experimental results (either in the supplemental material or as a URL)?
    \answerYes{[We include the code and instructions to reproduce the results here: \\ \url{https://github.com/vganapati/CT_PVAE}.]}
  \item Did you specify all the training details (e.g., data splits, hyperparameters, how they were chosen)?
    \answerYes{[Training details are available in the code documentation at \url{https://github.com/vganapati/CT_PVAE}.]}
        \item Did you report error bars (e.g., with respect to the random seed after running experiments multiple times)?
    \answerYes{[See Fig.~\ref{fig:foam_aggregate} in the Appendix.]}
        \item Did you include the total amount of compute and the type of resources used (e.g., type of GPUs, internal cluster, or cloud provider)?
    \answerYes{[See Section~\ref{sec:results}.]}
\end{enumerate}

\item If you are using existing assets (e.g., code, data, models) or curating/releasing new assets...
\begin{enumerate}
  \item If your work uses existing assets, did you cite the creators?
    \answerNA{}
  \item Did you mention the license of the assets?
    \answerNA{}
  \item Did you include any new assets either in the supplemental material or as a URL?
    \answerYes{[We release code for our developed framework.]}
  \item Did you discuss whether and how consent was obtained from people whose data you're using/curating?
    \answerNA{}
  \item Did you discuss whether the data you are using/curating contains personally identifiable information or offensive content?
    \answerNA{}
\end{enumerate}

\item If you used crowdsourcing or conducted research with human subjects...
\begin{enumerate}
  \item Did you include the full text of instructions given to participants and screenshots, if applicable?
     \answerNA{}
  \item Did you describe any potential participant risks, with links to Institutional Review Board (IRB) approvals, if applicable?
     \answerNA{}
  \item Did you include the estimated hourly wage paid to participants and the total amount spent on participant compensation?
    \answerNA{}
\end{enumerate}

\end{enumerate}


\appendix

\newpage

\section{Appendix}

\begin{figure}[h]
%

  \begin{subfigure}[t]{.45\textwidth}
    \centering
    \includegraphics[width=\linewidth]{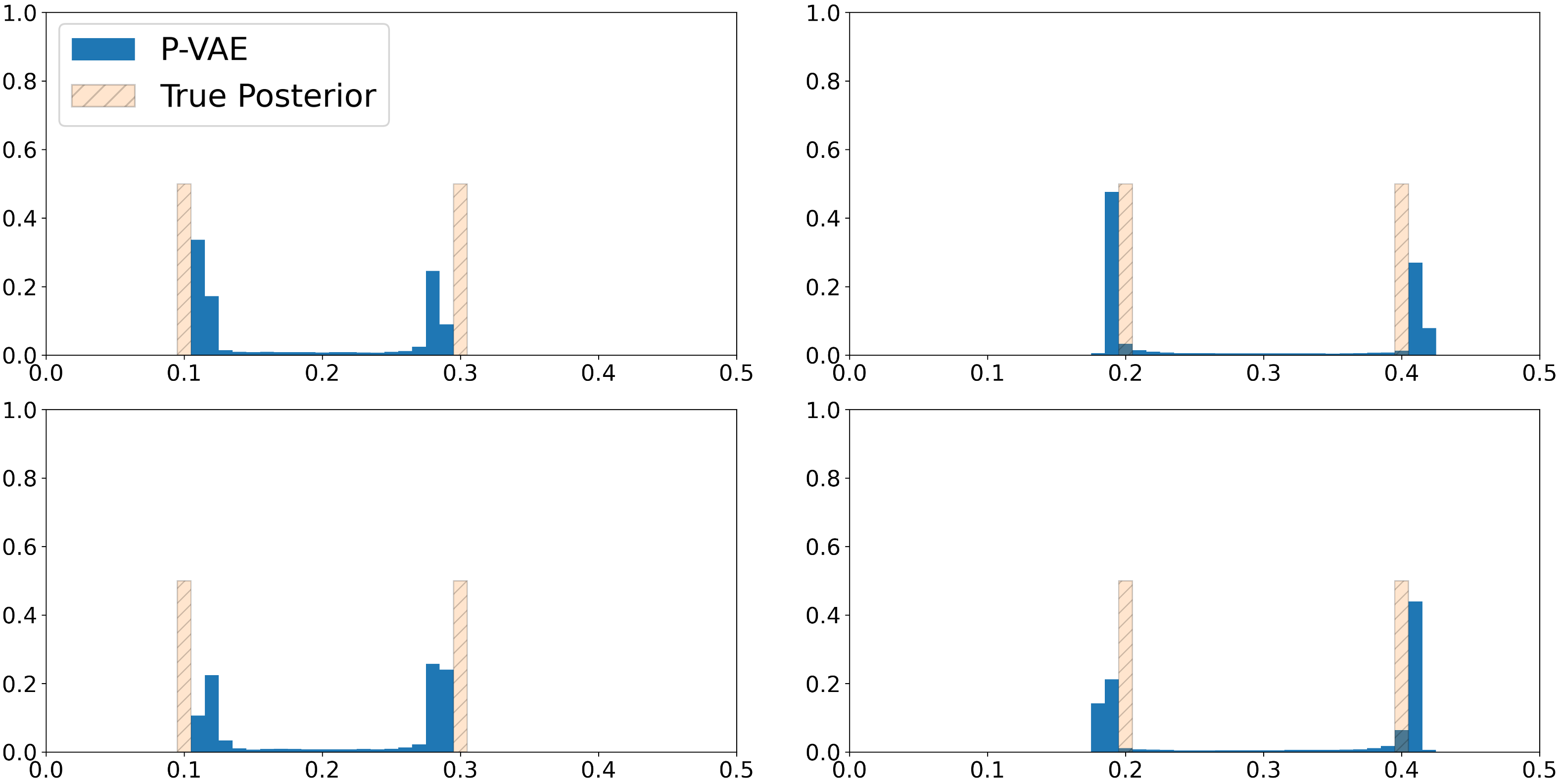}
    \caption{$O_2$ measured at $\frac{\pi}{2}$ radians.}
  \end{subfigure}
  \hfill
  \begin{subfigure}[t]{.45\textwidth}
    \centering
    \includegraphics[width=\linewidth]{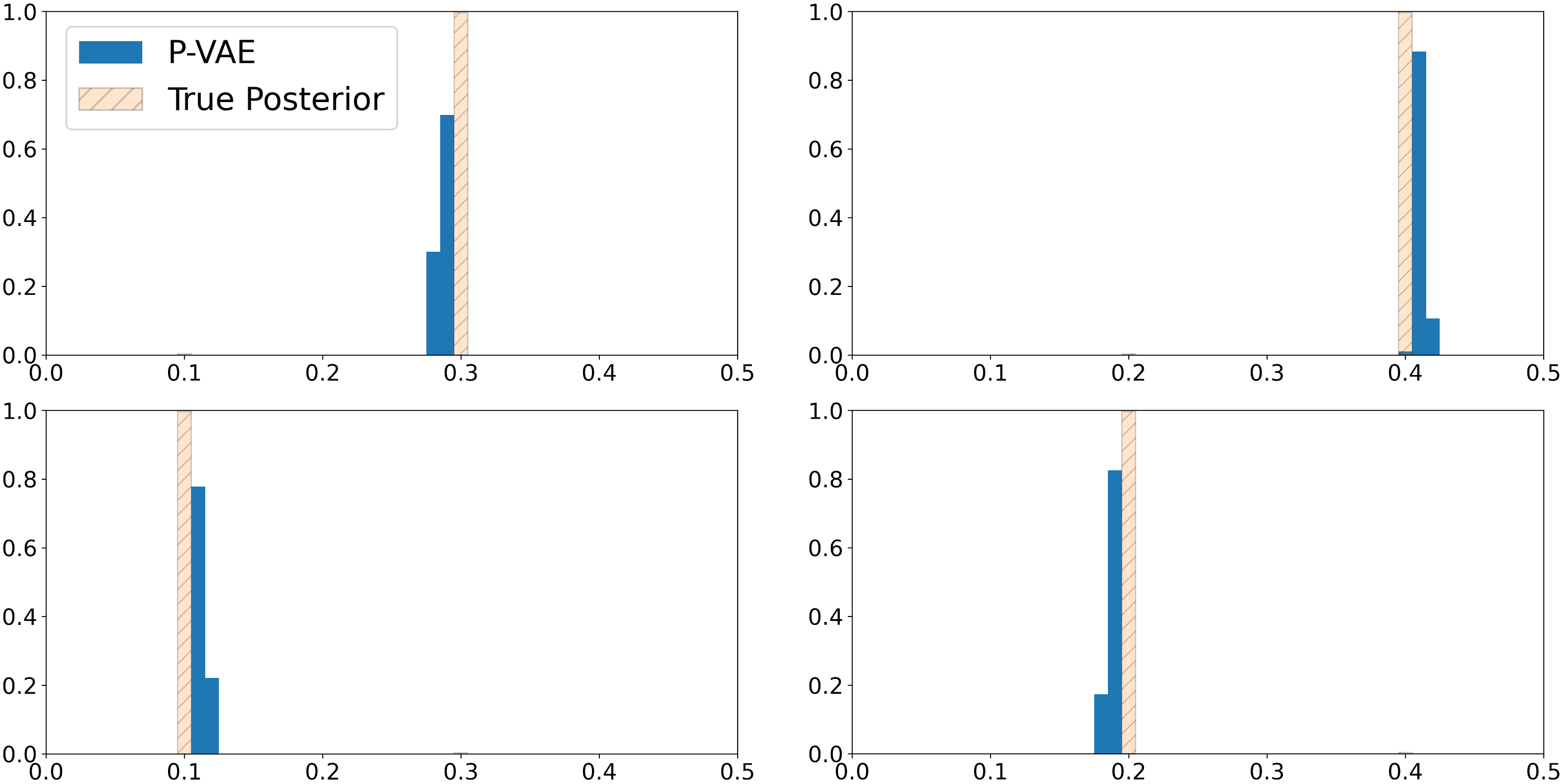}
    \caption{$O_2$ measured at $0$ radians.}
  \end{subfigure}
  \caption{Marginal posterior probabilities for each pixel from measurements on the toy dataset; measurements taken on object $O_2$ of the dataset.}
  \label{fig:toy_results2}
      \vspace{-1.0em}
\end{figure}

\begin{figure}[h]
\vspace{2em}
\centering
\includegraphics[width=1\textwidth]{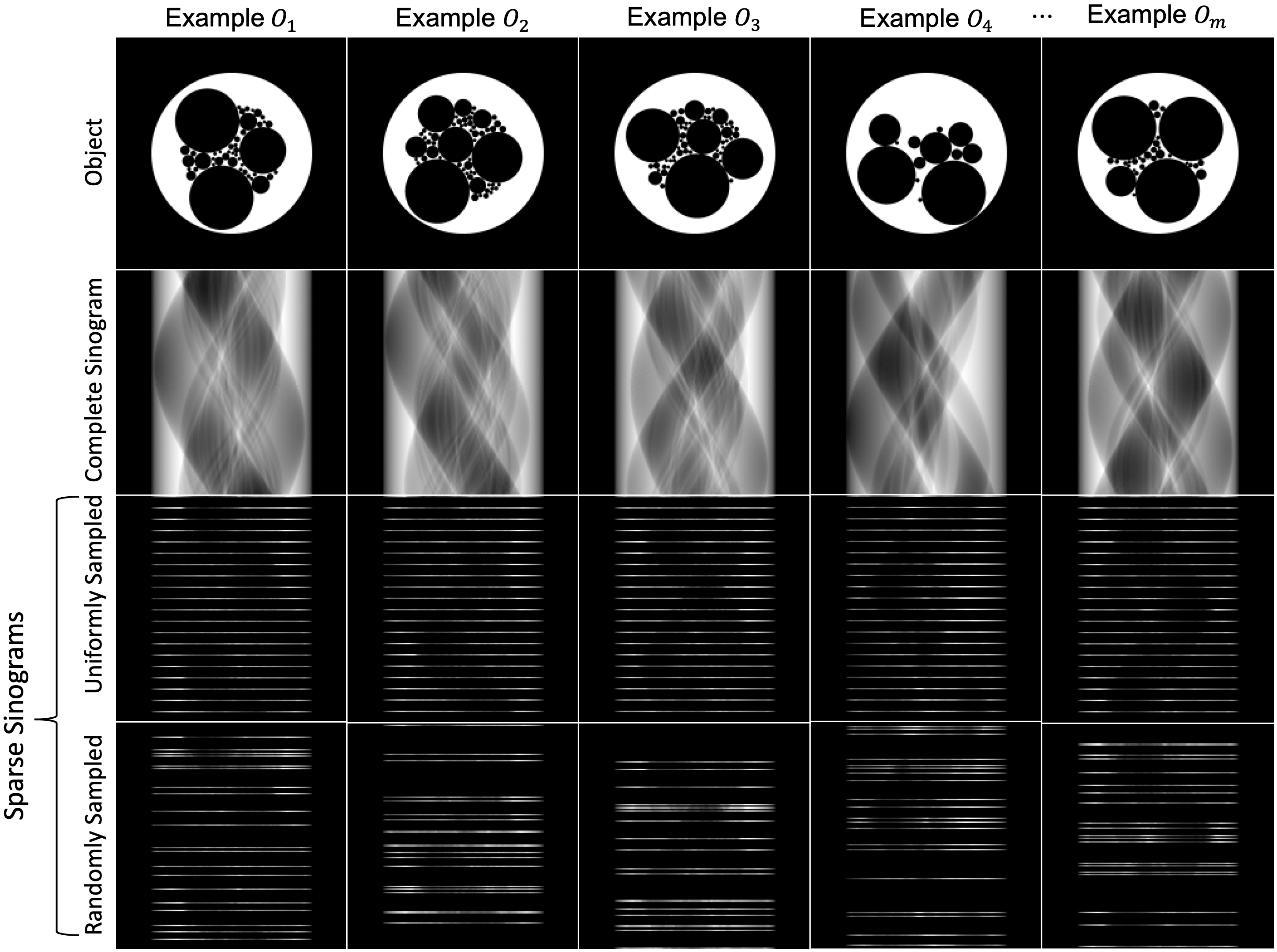}
\vspace{0em}
\caption{Object examples generated for the foam image dataset. Complete sinograms are taken at $180$ equally spaced rotation angles. Uniformly sparse sinograms are taken at $20$ equally spaced angles and randomly sampled sparse sinograms are taken at $20$ randomly chosen angles; these angles are re-sampled for each object of the dataset.}
\label{fig:foam}
\end{figure}

\begin{figure}[h]
\vspace{0em}
\centering
\includegraphics[width=1\textwidth]{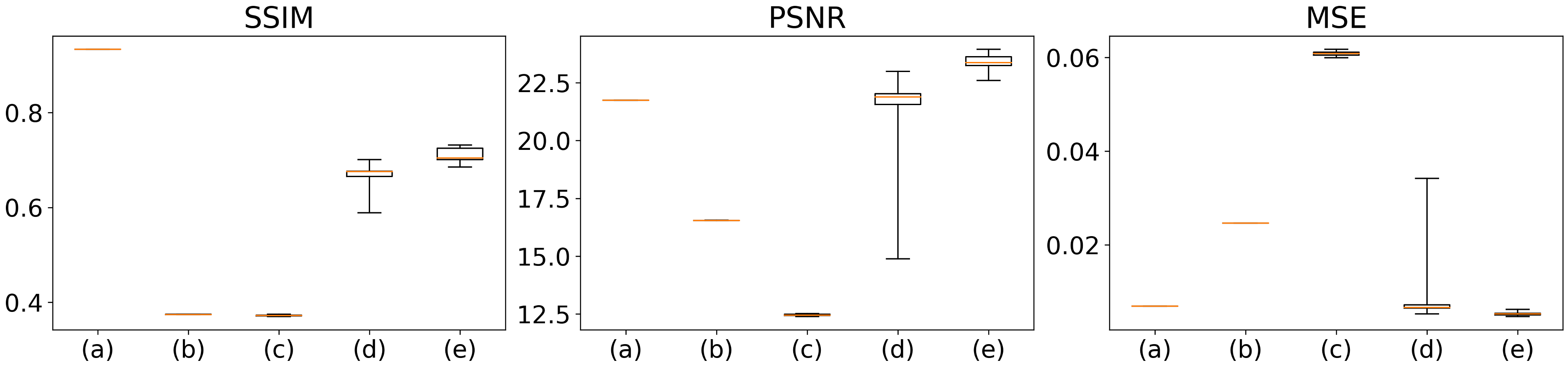}
\vspace{0em}
\caption{Image quality metrics of structural similarity (SSIM), peak signal-to-noise (PSNR), and mean-squared error (MSE) for reconstruction algorithms applied to the synthetic foam dataset, averaged over the entire dataset for 10 independent optimization trials. Reconstructions from (a) \texttt{gridrec} with the complete sinogram of $180$ rotation angles, (b) \texttt{gridrec} with the sparse sinogram of $20$ angles, uniformly spaced, (c) \texttt{gridrec} with the sparse sinogram of $20$ angles, randomly spaced, with each example having different angles chosen, (d) the physics-informed variational autoencoder with the sparse sinogram of $20$ angles, uniformly spaced, and (e) the physics-informed variational autoencoder with the sparse sinogram of $20$ angles, randomly spaced.}
\label{fig:foam_aggregate}
\end{figure}

\end{document}